\documentclass{article}

\pdfoutput=1
\usepackage{microtype}
\usepackage{graphicx}
\usepackage{subfig}
\usepackage{booktabs} % for professional tables
\usepackage{amssymb}
\usepackage{mathrsfs}
\usepackage{amsmath}
\usepackage{amsthm}
\usepackage{enumitem}
\usepackage{color}
\usepackage{hyperref}
\usepackage{natbib}
\usepackage{dsfont}
\usepackage{enumitem}
\usepackage{bbm}
\usepackage{algorithm}
\usepackage{algorithmic}

\newcommand\argmax{\mathop{\rm arg\,max}}

\newcommand {\defn} {\triangleq}
\newcommand \Reals {\ensuremath{\mathbb{R}}}
\newcommand \E {\mathop{\mbox{\ensuremath{\mathbb{E}}}}\nolimits}

\newcommand \pol {\pi}

\newcommand \bel {\xi}
\newcommand \mdp {\mu}
\newcommand \MDP {\mathcal{M}}
\newcommand \return {R}
\newcommand \utility {U}

\usepackage[accepted]{icml2019}

\icmltitlerunning{Epistemic Risk-Sensitive Reinforcement Learning}

\begin{document}

\twocolumn[
\icmltitle{Epistemic Risk-Sensitive Reinforcement Learning}




\begin{icmlauthorlist}
\icmlauthor{Hannes Eriksson}{c,z}
\icmlauthor{Christos Dimitrakakis}{c}
\end{icmlauthorlist}

\icmlaffiliation{c}{Department of Computer Science and Engineering, Chalmers University of Technology, Gothenburg, Sweden}
\icmlaffiliation{z}{Zenuity AB, Gothenburg, Sweden}

\icmlcorrespondingauthor{Hannes Eriksson}{hannese@chalmers.se}

\icmlkeywords{Machine Learning, ICML, Reinforcement Learning, Bayesian methods, Risk-aversion}

\vskip 0.3in
]



\printAffiliationsAndNotice{}  % leave blank if no need to mention equal contribution

\begin{abstract}
	We develop a framework for interacting with uncertain environments in reinforcement learning (RL) by leveraging preferences in the form of utility functions. We claim that there is value in considering different risk measures during learning. In this framework, the preference for risk can be tuned by variation of the parameter $\beta$ and the resulting behavior can be risk-averse, risk-neutral or risk-taking depending on the parameter choice. We evaluate our framework for learning problems with model uncertainty. We measure and control for \emph{epistemic} risk using dynamic programming (DP) and policy gradient-based algorithms. The risk-averse behavior is then compared with the behavior of the optimal risk-neutral policy in environments with epistemic risk.
\end{abstract}

	\section{Introduction}
In this work, we consider the problem of reinforcement learning (RL) for risk-sensitive policies under epistemic uncertainty, i.e. the uncertainty due to the agent not knowing how the environment responds to the agent's actions. This is in contrast to typical approaches in risk-sensitive decision making, which have focused on the aleatory uncertainty due to inherent stochasticity in the environment. Modelling risk-sensitivity in this way makes more sense for applications such as autonomous driving, where systems are nearly deterministic, and most uncertainty is due to the lack of information about the environment. In this paper, we consider epistemic uncertainty and risk sensitivity both within a Bayesian utilitarian framework. We develop novel algorithms for policy optimisation in this setting, and compare their performance quantitatively with policies optimised for the \emph{risk-neutral} setting.

%We develop novel algorithms for policy optimisation in this setting, and compare their performance quantitatively with policies optimised with respect to aleatory risk. 

Reinforcement learning (RL) is a sequential decision-making problem under uncertainty. Typically, this is formulated as the maximisation of an expected return, where the return $\return$ is defined as the sum of scalar rewards obtained over time $\return \defn \sum_{t=1}^T r_t$ to some potentially infinite or random horizon $T$. As is common in RL, we assume that agents acting in a discrete Markov decision process (MDP). For any given finite MDP $\mdp \in \MDP$, the optimal risk-neutral policy $\pol^*(\mdp) \in \argmax_\pol \E_\mdp^\pol[\return]$  be found efficiently via dynamic programming algorithms. Because the learning problem specifies that $\mdp$ is unknown, the optimal policy under epistemic uncertainty must take into account expected information gain. In the Bayesian setting, we maintain a subjective belief in the form of a probability distribution  $\bel$ over MDPs $\MDP$ and the optimal solution is given by:

\begin{equation}
\label{eq:bayes-maximisation}
\max_\pol \E_\bel^\pol[\return] = \max_\pol \int_\MDP \E_\mdp^\pol[\return] \, d \bel(\mdp).
\end{equation}
This problem is generally intractable, as the optimisation is performed over  \emph{adaptive} policies, which is exponentially large in the problem horizon. A risk-neutral agent would only wish to maximise expected return. However, we are interested in the case where the agent is risk-sensitive, and in particular with respect to uncertainty about $\mdp$.

\paragraph{Contribution.} Typically, risk sensitivity in reinforcement learning has  addressed risk due to the inherent environment stochasticity (aleatory) and that due to uncertainty (epistemic) with different mechanisms. We instead wish to consider both under a coherent utility maximising framework, where the convexity of the utility with respect to the return $\return$ determines whether our behaviour will be risk-seeking or risk-averse. By applying this utility function on the actual return or the expected return given the MDP we can easily trade off between aleatory and epistemic risk-sensitivity.

\subsection{Related work}
Defining risk sensitivity with respect to the return $\return$ can be done in a number of ways. In this paper, we focus on the expected utility formulation, whereby the utility is defined as a function of the return: concave functions lead to risk aversion and convex to risk seeking. In the context of reinforcement learning, this approach was first proposed by~\citet{mihatsch2002risk}, who derived efficient temporal-difference algorithms for exponential utility functions. However, the authors considered risk only with respect to the MDP stochasticity.

Works on Robust MDPs have traditionally dealt with uncertainty due to epistemic risk. In~\citet{givan2000bounded} the authors extend the exact MDP setting by allowing bounded transition probabilities and rewards. They introduce optimal pessimistic and optimistic policy updates for this setting.
\citet{mannor2012lightning} further extend this by allowing the parameter uncertainties to be coupled which allows for less conservative solutions.
\citet{tamar2014scaling} use RL and approximate dynamic programming (ADP) to scale the Robust MDP framework to problems with larger state spaces using function approximators.

Another way of dealing with risk is conditional value-at-risk (CVaR)~\cite{rockafellar2000optimization, sivara2017} measures. Compared to more traditional risk measures such as Mean-Variance trade-off~\cite{markowitz1952portfolio, tamar2012policy} or expected utility framework~\cite{friedman1948uac}, CVaR allows us to control for tail risk.
CVaR has been used for risk-sensitive MDPs in \citet{chow2014algorithms,chow2015risk,chow2015riskconstrained,stanko2018distribuvcni}.

A Bayesian setting is also considered by~\citet{depeweg2018decomposition}, who focus on a risk decomposition in aleatory and epistemic components. The authors model the underlying dynamics with a neural network. Under the risk-sensitive framework, they aim to trade-off expected performance with variation in performance. The risk-sensitive criterion they aim to maximise is $\mathbb{E}\big[\sum_t r_t\big] + \beta\sigma\big(\sum_t r_t\big)$ where $\beta$ controls for the level of risk-aversion. Thus, they are essentially considering risk only with respect to individual rewards. Our paper instead considers the risk with respect to the total return, which we believe is a more interesting setting for long-term planning problems under uncertainty. 

\section{Optimal policies for epistemic risk}
\label{sec:epistemic_setting}

%\cdcomment{Introduce the problem, and how we are modelling it.}

Under the expected utility hypothesis, risk sensitivity can be modelled~\citep{friedman1948uac} through a concave or convex utility function $\utility : \Reals \to \Reals$ of the return $\return$. Then, for a given model $\mdp$, the optimal $\utility$-sensitive policy with respect to \emph{aleatory} risk is the solution to $\max_\pol \E_\mdp^\pol [\utility(\return)]$. In the case where we are uncertain about what is the true MDP, we can express it through belief $\bel$ over models $\mdp$. Then optimal policy is the solution to
\begin{equation}
\pol^A(\utility, \bel) \defn \argmax_\pol \int_\MDP \E_\mdp^\pol [\utility(\return)] \, d \bel(\mdp).
\label{eq:aleatory-bayes-risk}
\end{equation}
However, this is only risk-sensitive with respect to the stochasticity of the underlying MDPs. We believe that \emph{epistemic} risk, i.e. the risk due to our uncertainty about the model is more pertinent for reinforcement learning. The optimal \emph{epistemic risk sensitive} policy maximises:
\begin{equation}
\pol^E(\utility, \bel) \defn \argmax_\pol \int_\MDP \utility\left(\E_\bel^\pol (\return)\right) \, d \bel(\mdp).
\label{eq:epistemic-bayes-risk}
\end{equation}
When $\utility$ is the identity function, both solutions are risk-neutral. In this paper, we shall consider functions of the form $\utility(x) = \beta^{-1}e^{\beta x}$, so that $\beta > 0$ is risk-seeking and $\beta < 0$ risk-averse.

We consider two algorithms for this problem. The first, based on an approximate dynamic programming algorithm for Bayesian reinforcement learning introduced in~\cite{dimitrakakis:mmbi:ewrl:2011}, is introduced in Section~\ref{sec:adp}. 
The second, based on the policy gradient framework~\cite{sutton2000policy}, allows us to extend the previous algorithm to larger MDPs and allows for learning of stochastic policies. The priors used in the experiments are detailed in Section~\ref{sec:experiment}.

\subsection{Utility functions.}
\label{sec:utility}
Previous works~\cite{mihatsch2002risk,howard1972risk} on utility functions for risk-sensitive reinforcement learning used an \emph{exponential} utility function of the form $U(x) = \beta^{-1} e^{\beta x}$. The $\beta$ parameter could then be used to control the shape of the utility function and determine how risk-sensitive we want to be with respect to $x$. This is the utility function we will use in this paper.\footnote{Other choices are $U(x) = x^\beta$ or $U(x) = \log(x)$, however they both come with significant drawbacks, such as handling negative returns $R$.}
\citet{mihatsch2002risk} consider not only exponential utility functions, but a special form of them, that is:
\begin{equation}
\pi^E(U) = \underset{\pol}{\arg\max} \frac{1}{\beta} \log{} \mathbb{E}\Big(\exp(\beta R)\Big)
\end{equation}
\citet{mihatsch2002risk,coraluppi1999risk,marcus1997risk} argue that this utility function has some interesting properties. In particular, maximising leads to maximising:
\begin{equation}
\mathbb{E}[R] + \frac{\beta}{2}Var[R] + \mathcal{O}(\beta^2)
\end{equation}
From this, we get that the behavior of our policy $\pol^E(U)$ as $\beta \rightarrow 0$ is the same as for the risk-neutral case. For $\beta < 0$ we get risk-averse behavior and for $\beta > 0$ risk-taking behavior. 

In our work, we are working in a Bayesian setting. Consequently, we introduce a belief $\bel$ over models and define the expected utility of a policy $\pi$ related to the model uncertainty as follows:
\begin{equation}
\label{eq:exp-utility}
U^E_\beta(\xi, \pi) \defn \frac{1}{\beta}\log\int_{\mathcal{M}} \exp\big(\beta\mathbb{E}_\mdp^\pol[R]\big)d\xi(\mu).
\end{equation}

\subsection{Epistemic risk sensitive backward induction}
\label{sec:adp}

Algorithm~\ref{alg:evi} is an Approximate Dynamic Programming (ADP)~\cite{powell2007approximate} algorithm for optimising policies in our setting. While the algorithm is given for a belief over a finite set of MDPs, it can be easily extended to arbitrary $\bel$ through simple Monte-Carlo sampling, as in~\citet{dimitrakakis:mmbi:ewrl:2011}.

The algorithm essentially maintains a separate $Q_\mdp$-value function for every MDP $\mdp$. At every step, it finds the best local policy $\pol$ with respect to the utility function $\utility$. Then the value function $V_\mdp$ of each MDP reflects the value of $\pol$ within that MDP.

%\subsection{Eventual risk-neutrality}
%If the Bayesian posterior converges, then the behaviour of the algorithm becomes risk neutral.
%\cdcomment{What exactly do we want to show here and why?}
%\hecomment{We no longer need to say anything about it since we always have epistemic risk in the experiments right now}
%\begin{lemma}
%	\label{lemma:belief_mdp}
%	If $\mdp^*$ is a fixed MDP such that $\bel_t \to \delta(\mdp^*)$ then the limiting epistemic policy is risk-neutral.
%\end{lemma}
%\begin{proof}
%  When $\bel = \delta(\mdp^*)$ then 
%  \begin{align}
%    \underset{\pol}{\max{}} \int_\MDP U(V_{\mdp^*}^\pol)\bel(\mdp)d\mdp
%    &= 
%      \underset{\pol}{\max{}} U(V_{\mdp^*}^\pol) \int_\MDP \bel(\mdp)d\mdp
%      \\
%      &=
%      \underset{\pol}{\max{}} U(V_{\mdp^*}^\pol)
%  \end{align}
%\end{proof}

This algorithm will be used as a baseline for the experiments with risk aversion. %From Lemma~\ref{lemma:belief_mdp} 
We know that as the epistemic uncertainty vanishes; if there is only one underlying true model, then the optimal policy for the epistemic risk-sensitive algorithm will be the same as the optimal policy for the epistemic risk-neutral algorithm. It is important to point out that this is only true when our belief is concentrated around the true MDP. Behavior during learning of the MDP could be very different. For cases with multiple models, the epistemic uncertainty will always exist and there are no such guarantees.

\begin{algorithm}[tb]
	\caption{Epistemic Risk Sensitive Backwards Induction}
	\label{alg:evi}
	\begin{algorithmic}
		\STATE {\bfseries Input:} $\mathcal{M}$ (set of MDPs), $\bel$ (current posterior)
		\REPEAT
		\FOR{$\mdp \in \mathcal{M}$ $s \in \mathcal{S}$, $a \in \mathcal{A}$}
		\STATE $Q_\mdp(s,a) = \mathcal{R}_\mdp(s,a) + \gamma \sum_{s'} \mathcal{T}_\mdp^{ss'} V_\mdp(s')$
		\ENDFOR
		\FOR{$s \in \mathcal{S}$}
		\FOR{$a \in \mathcal{A}$}
		\STATE $\mathcal{Q}_\bel(s,a) = \sum_\mdp \bel(\mdp) \utility[(Q_\mdp(s,a)]$
		\ENDFOR
		\STATE $\pol(s) = \argmax_a \mathcal{Q}_\bel(s,a)$.
		\FOR{$\mdp \in \mathcal{M}$}
		\STATE $V_\mdp(s) = {Q}_\mdp(s,\pol(s))$.
		\ENDFOR
		\ENDFOR
		\UNTIL{\itshape convergence}
		\STATE {\bfseries return $\pol$}
	\end{algorithmic}
\end{algorithm}

\subsection{Bayesian policy gradient}
\label{sec:bpg}
A common method for model-free reinforcement learning is policy gradient~\cite{sutton2000policy}. It is also very useful in model-based settings, and specifically for the Bayesian reinforcement learning problem, where sampling from the posterior allows us to construct efficient stochastic gradient algorithms. Bayesian policy gradient (BPG) methods have been explored for these contexts before,~\citet{ghavamzadeh:bpga} uses BPG to plan how to maximise information gain given our current belief $\xi_t$ for the risk-neutral setting. In our specific setting, we are interested in maximising (\ref{eq:epistemic-bayes-risk}), where we use utility function~(\ref{eq:exp-utility}) to model risk sensitivity. 

Our choice of policy parametrisation is a softmax policy with non-linear features. The probability of selecting action $a$ in state $s$, given current parameters $\theta$, is $\pi_\theta(a|s) = \frac{e^{\phi(s,a, \theta)}}{\sum_{a' \in \mathcal{A}}e^{\phi(s,a',\theta)}}$, where the features $\phi(s, a, \theta)$ are calculated by a feedforward neural network with one hidden layer. Full parameter details of the neural networks are given in Section~\ref{sec:experiment}.

We choose to introduce a policy parametrisation over (\ref{eq:epistemic-bayes-risk}). This gives us (\ref{eq:epistemic-bayes-risk-theta}).
\begin{equation}
\label{eq:epistemic-bayes-risk-theta}
\pi^E(U, \xi, \theta) = \int_{\mathcal{M}} U\Big(\mathbb{E}_{\pi_\theta}^\mu[R]\Big)d\xi(\mu)
\end{equation}

Combining our choice of utility function in (\ref{eq:exp-utility}) and (\ref{eq:epistemic-bayes-risk-theta}) gives us our new objective function (\ref{eq:epistemic-bayes-risk-theta-utility}). 

\begin{equation}
\label{eq:epistemic-bayes-risk-theta-utility}
\pi^E(U, \xi, \theta) = \frac{1}{\beta} \log{} \int_{\mathcal{M}} \exp\Big(\beta\mathbb{E}_{\pi_\theta}^\mu[R]\Big)d\xi(\mu)
\end{equation}

Our goal is now to find the set of parameters $\theta$ so as to maximise information gain. Taking the gradient of (\ref{eq:epistemic-bayes-risk-theta-utility}) gives (\ref{eq:pgbefore}) which leads to (\ref{eq:pg}) after a straightforward derivation.

\begin{equation}
\label{eq:pgbefore}
\nabla_\theta \pi^E(U, \xi, \theta) =\nabla_\theta \frac{1}{\beta} \log{} \int_{\mathcal{M}} \exp\Big(\beta\mathbb{E}_{\pi_\theta}^\mu[R]\Big)d\xi(\mu)
\end{equation}

\begin{equation}
\label{eq:pg}
=\frac{\int_\mathcal{M}\exp\Big(\beta \mathbb{E}_{\pi_\theta}^\mu[R]\Big)\nabla_\theta\mathbb{E}_{\pi_\theta}^\mu[R]d\xi(\mu)}{\int_{\mathcal{M}} \exp\Big(\beta\mathbb{E}_{\pi_\theta}^\mu[R]\Big)d\xi(\mu)}
\end{equation}

The RHS of the numerator is the classical policy gradient~\cite{sutton2000policy}. We can replace the integrals in~(\ref{eq:pg}) with a sum by sampling models from our belief $\xi$. Note that this has to be done independently for the numerator and the denominator to avoid bias. To get each of the separate $\mathbb{E}_{\pol_\theta}^{\mu}[R]$ we make use of rollouts. Each expected return term is estimated independently with their own set of rollouts. 

We now have an estimate for the gradient and can move our policy parameters accordingly to act optimally given our belief $\xi_t$.

The procedure of calculating the gradient in ({\ref{eq:pg}}) is given in Algorithm~\ref{alg:papg}. The algorithm builds upon the classic episodical \emph{REINFORCE} algorithm~\cite{williams1992simple}. Basing the algorithm on an episodical update makes sense in this case since it decreases the number of rollouts we have to do. One drawback is that it restricts updates episode by episode and so the information gained in the current episode cannot be used until after the episode ends. Monte-Carlo methods such as \emph{REINFORCE} also have a few other drawbacks such as high variance and low convergence, problems that could be addressed by extending the current framework to an actor-critic based one, as in~\citet{barto1983neuronlike,ghavamzadeh2007bayesian,ghavamzadeh2016bayesian}.

%Grammaly identified multiple weird sentences here so should rewrite it...
The algorithm consists of two stages. One planning stage which is used to identify the best policy parameters $\theta_t$ given our current belief $\xi_t$. This is done through model-based simulation by sampling MDPs from our belief. After the rollouts have been collected the neural network is optimised and a new policy $\pi_{\theta_{t+1}}$ is attained. We then commit to this policy for one episode and move on to the next stage of the algorithm.

The second stage of the algorithm uses the new policy to act in the real environment. Trajectories $\tau = (s_0, a_0, r_1, s_1, ..., s_{T-1}, a_{T-1}, r_T, s_T)$ are collected from this environment and used to update our belief over transitions and reward functions.

\begin{algorithm}[tb]
	\caption{Epistemic Risk Sensitive Policy Gradient}
	\label{alg:papg}
	\begin{algorithmic}
		\STATE {\bfseries Input:} Policy parametrisation $\theta_t$, $\bel_t$ (current posterior).
		\REPEAT
		\STATE $\texttt{Simulate to get }\theta_{t+1}$
		\FOR{$i=1$ {\bfseries to} $N$}
		\STATE $\mdp_{(1)}, \mdp_{(2)} \sim (\mathcal{M}_t, \mathcal{R}_t)$
		\FOR{$j=1$ {\bfseries to} $M$}
		\STATE $\tau_{\mdp_{(1)}}^{(1)}, \tau_{\mdp_{(1)}}^{(2)} \sim \pol_\theta, \mdp_{(1)}$
		\STATE $\tau_{\mdp_{(2)}}^{(3)} \sim \pol_\theta, \mdp_{(2)}$
		\ENDFOR
		\ENDFOR
		\STATE $\theta_{t+1} \leftarrow \theta_t - \frac{\sum_{i=0}^N \exp\Big(\beta {\tau_{\mdp_i}}^{(1)}\Big){\tau_{\mdp_i}}^{(2)}\nabla_\theta\log{}\pi_\theta(a|s)}{\sum_{i=0}^N \exp\Big(\beta {\tau_{\mdp_i}}^{(3)}\Big)}$
		\STATE $\texttt{Deploy }\pol_{\theta_{t+1}}$
		\STATE $\tau \sim \mdp, \pol_{\theta_{t+1}}$
		\STATE $\bel_{t+1} \leftarrow \bel_t, \tau$
		\UNTIL{\itshape convergence}
	\end{algorithmic}
\end{algorithm}

\subsection{Bayesian Epistemic CVaR}
\label{sec:risk-sensitive}

Another common approach of handling risk-aversion is to optimize with respect to a CVaR objective, \cite{chow2014algorithms, chow2015risk, tamar2015optimizing, stanko2018distribuvcni}. In this paper, we investigate Bayesian epistemic CVaR, first studied by~\citet{chen2018leverage} in the context of investment strategies. In their case, they use it to do posterior inference of a SV-ALD model to efficiently estimate risk. We define Bayesian epistemic CVaR as follows, defining the set of MDPs where we get at least $z$ utility as the following;

\begin{equation}
\MDP_z^\pi \defn \Big\{ \mdp \, | \, \E_\mdp^\pol[R] \geq z \Big\}.
\end{equation}

The $\nu_\bel^\pol(\alpha)$ is value-at-risk in the Bayesian setting for a given quantile $\alpha$. 

\begin{equation}
\nu^\pol_\bel(\alpha) = \inf \Big\{z \, | \, \bel(\MDP_z^\pol) \geq \alpha\Big\}.
\end{equation}

\begin{align}
C^\pol_\bel(\alpha)
&= \E^\pol_\bel\Big[ \E_\mdp^\pol[R] \mid \E_\mdp^\pol[R] \leq \nu^\pol_\bel(\alpha)\Big]
\\
&= \int_{\MDP^\pol_{\nu^\pol_\bel(\alpha)}} \E_\mdp^\pol[R] \, d \bel(\mdp).
\end{align}

%\cdcomment{So then getting the gradient for $C$ should basically mean that we sample from the set  $\MDP^\pol_z$ and doing a gradient step, i.e. ignoring the MDPs which are above the quantile for the current policy. Of course, an MDP could suddenly go into the quantile after we change our policy.}

Concisely stated; we want to maximize our performance for the $\alpha$ least likely MDPs according to our belief $\bel$. The intuition behind this is that we want to be risk-averse with respect to what we are the most uncertain about.

\section{Experiments}
\label{sec:experiment}

In this paper, we conduct two kinds of experiments. Firstly, a classical learning problem on a \emph{Gridworld}. This is discrete, so we can use Algorithm~\ref{alg:evi} to find a near-optimal deterministic policy. Similar problems have been studied before in the field of robust MDPs and there are already solutions for finite state and action-space problems. 
As this is a discrete state-action space, the belief maintained over the MDP transitions is in the form of a Dirichlet-product prior and Belief over rewards is in the form of a NormalGamma prior. Further experimental details about this problem can be found in Section~\ref{ex:gw} and Appendix~\ref{sec:agw}. 

Secondly, we consider a continuous state-space problem in Section~\ref{ex:opt}. We do not see a trivial extension of previous works in robust MDPs to the case with infinite state-space. Previous works such as~\citet{tamar2014scaling} scale earlier works on robust MDP to large state-space problems but not to continuous. The problem has been studied to a great extent in the field of risk-aversion, see~\citet{tamar2012policy, tamar2015optimizing, chow2014algorithms} and Appendix~\ref{sec:ao}. For this problem, we maintain function priors in the form of Gaussian Processes on the reward functions and transition kernels.

%\cdcomment{What is the point of the experiments? Perhaps we want to show that it is possible to get better / safer exploration by being risk sensitive. However, we normally assume that the utility function (as well as the reward function) are somehow externally imposed, and hence do not need a justification of their own. It is simply that the decision maker is interested in the ability to use a risk-sensitive policy. This means we should show at least: (a) That the algorithm(s) correctly maximise(s) the utility we care about efficiently (b) That one algorithm is better in achieving the objective (c) that being risk-sensitive allows in practice to achieve a different return distribution: high-risk should mean we are able to more often find optimal policies, but also we often get stuck - low-risk that we almost always find a good sub-optimal policy. In those cases we should really measure the return as given wrt the discount factor we have set. In addition, we can look at experimental results for average reward over time, even though this is not what we care about, to show that speed of exploration and asymptotic performance depend greatly on the choice of beta.}

\subsection{Gridworld experiment}
\label{ex:gw}

\begin{figure}
	\caption{Experiment detailing the results of the run of \emph{ADP} Algorithm~\ref{alg:evi} for varying choices of $\beta$. (a) The top plot of the grid is the regret over time with respect to the optimal deterministic risk-neutral policy. (b) The bottom plot shows the distribution of falls throughout over all experiments.}
	\includegraphics[width=\linewidth]{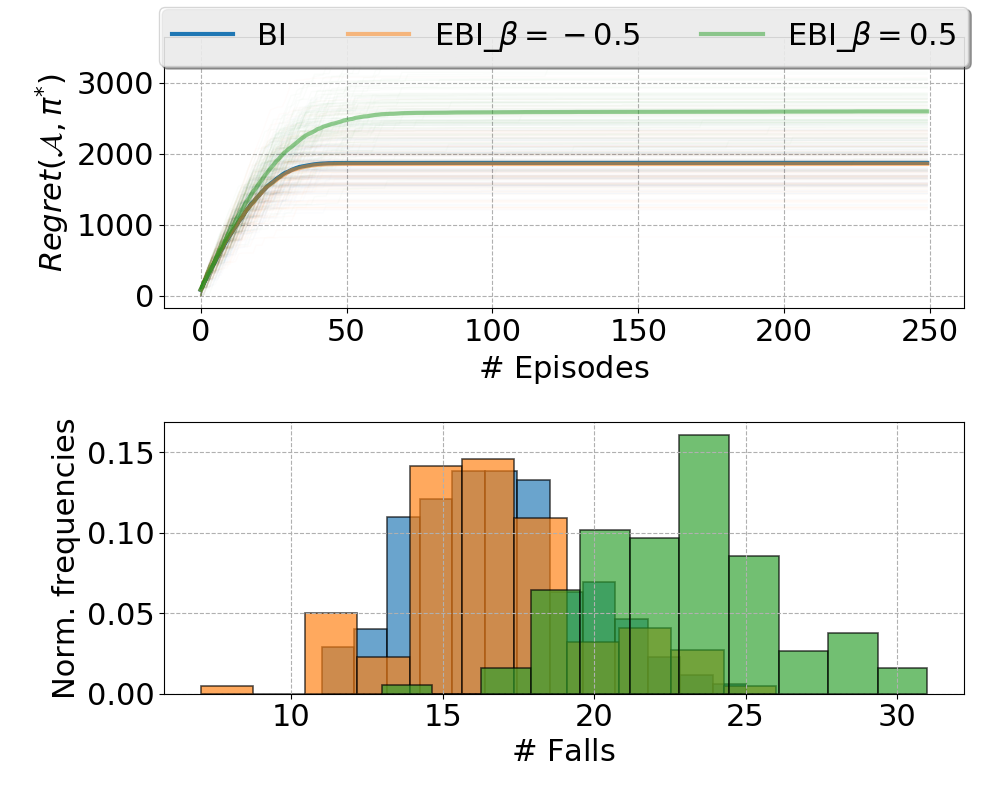}
	\centering
	\label{fig:gw}
\end{figure}

Shown in Figure~\ref{fig:gw} are the results of Algorithm~\ref{alg:evi} ran for different values of $\beta$. We aimed to test both risk-aversion and risk-taking behavior through the choice of this parameter. The regret is averaged over $\approx 100$ independent experiment runs.  We can see that the regret increases for more risk-taking policies in the top plot. In this environment the risk-neutral and the risk-averse behavior is almost identical, with only a minor difference in \emph{$\# Falls$}. The risk-taking policy is in one sense, over-exploring while the risk-averse policy in general, would be more inclined to do exploitation. Note that in this experiment the only epistemic uncertainty comes from that the agent does not know the true MDP $\mu$. This uncertainty will go down over time as more transitions are observed.

\subsection{Option pricing experiment}
\label{ex:opt}

\begin{figure}
	\caption{Experiment detailing the results of the run of Epistemic Bayesian Policy gradient \emph{EBPG} Algorithm~\ref{alg:papg} for varying choices of $\beta$. For comparison we also include a risk-neutral \emph{PG}, a risk-neutral \emph{BPG} and \emph{CVaR BPG}. (a) The upper plot is a histogram over returns for all episodes observed. (b) The bottom plot is a histogram of the return over the last 10000 episodes, so the most current policy. }
	\includegraphics[width=\linewidth]{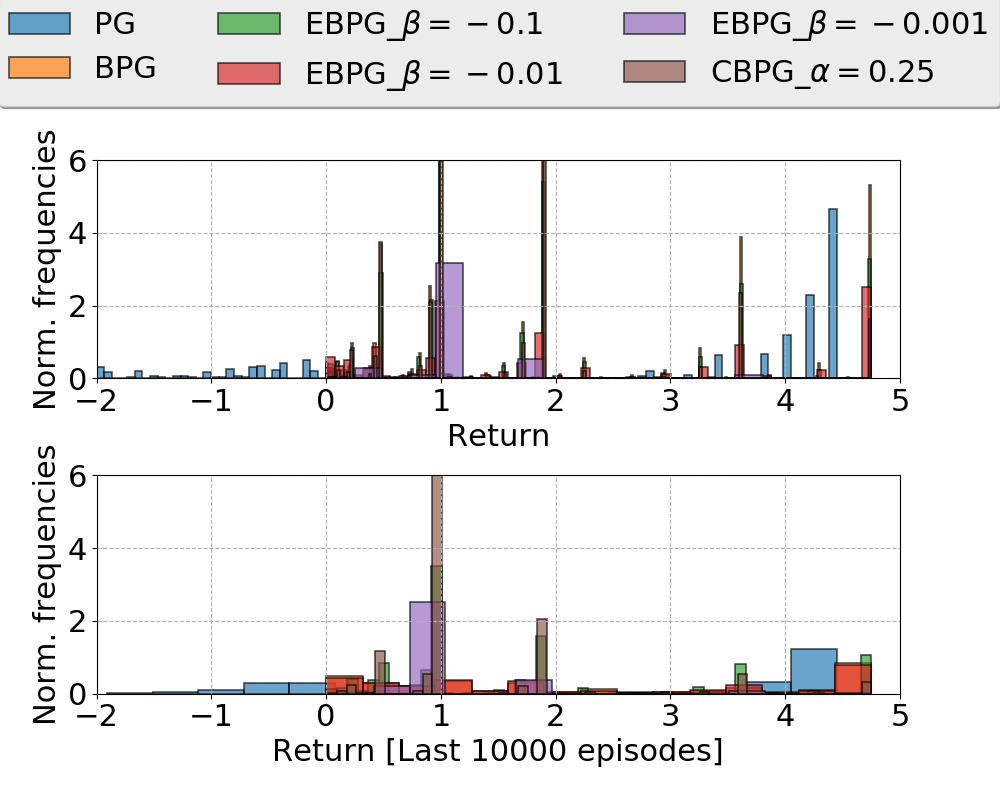}
	\centering
	\label{fig:opt}
\end{figure}

The result of runs in this environment is depicted in Figure~\ref{fig:opt}. The blue colored line will serve as a reference and has been verified to have an almost optimal risk-neutral policy. It also has the most observed data (millions of episodes) compared to the other algorithms (hundreds of thousands of episodes). We do see some notable difference in behavior between the policies and can identify that the algorithm with CVaR objective and the algorithm with $\beta=-0.001$ are overly cautious. On the other hand for $\beta=\{-0.01, -0.1\}$ and BPG we see behavior similar to that of regular PG.

\section{Conclusions}
We have introduced a framework that allows us to control for how risk-sensitive we want to be with respect to model uncertainty in the Bayesian RL setting. In Section~\ref{ex:gw} we detail the performance of Algorithm~\ref{alg:evi}, an ADP algorithm that has the ability to control for risk-sensitiveness in environments with discrete state and action-spaces. The results point towards risk-averse policies lead to less exploratory policies; that is, we would only expect to explore additionally if we are convinced there is much more to be gained by doing so, compared to a risk-neutral policy. Figure~\ref{fig:opt} shows an extension of this, using Policy gradient algorithms to more complicated environments with continuous state-space. 

A few hurdles noted is the speed of Algorithm~\ref{alg:papg}. In order to get a rich representation of the model uncertainty, a lot of MDPs should be sampled. However, sampling MDPs in this setup is quite expensive. Other belief models could be considered instead of GP if we were to scale this up to real problems.

\paragraph{Future work.} 
We can see an adaptive agent that could change its risk-sensitive parameter $\beta$ with time. For problems where we know uncertainties will resolve later into the future, this could be one approach.

There could be an avenue to explore using utility functions to induce more exploratory behavior in complicated environments. Current methods use concepts such as curiosity~\cite{houthooft2016vime,pathak2017curiosity} for reward shaping or entropy regularization~\cite{williams1991function, mnih2016asynchronous,o2016combining} to enforce additional exploration. However, under the expected utility framework we could perhaps get this for free by changing the utility function.

Decreasing variance and speeding up Algorithm~\ref{alg:papg} is of utmost importance for this framework to be used for more complex problems. As we touched upon in Section~\ref{sec:bpg} a straight-forward improvement would be to move towards the Bayesian actor-critic framework introduced in~\citet{ghavamzadeh2007bayesian}. 

\section*{Acknowledgement}
This work was partially supported by the Wallenberg AI, Autonomous Systems and Software Program (WASP) funded by the Knut and Alice Wallenberg Foundation.\\
The simulations were performed on resources
provided by the Swedish National Infrastructure for Computing (SNIC)
at Chalmers Centre for Computational Science and Engineering (C3SE).


\bibliography{references}

\begin{thebibliography}{34}
\providecommand{\natexlab}[1]{#1}
\providecommand{\url}[1]{\texttt{#1}}
\expandafter\ifx\csname urlstyle\endcsname\relax
  \providecommand{\doi}[1]{doi: #1}\else
  \providecommand{\doi}{doi: \begingroup \urlstyle{rm}\Url}\fi

\bibitem[Barto et~al.(1983)Barto, Sutton, and Anderson]{barto1983neuronlike}
Barto, A.~G., Sutton, R.~S., and Anderson, C.~W.
\newblock Neuronlike adaptive elements that can solve difficult learning
  control problems.
\newblock \emph{IEEE transactions on systems, man, and cybernetics}, \penalty0
  (5):\penalty0 834--846, 1983.

\bibitem[Chen et~al.(2018)Chen, Zerilli, and Baum]{chen2018leverage}
Chen, L., Zerilli, P., and Baum, C.~F.
\newblock Leverage effects and stochastic volatility in spot oil returns: A
  bayesian approach with var and cvar applications.
\newblock \emph{Energy Economics}, 2018.

\bibitem[Chow \& Ghavamzadeh(2014)Chow and Ghavamzadeh]{chow2014algorithms}
Chow, Y. and Ghavamzadeh, M.
\newblock Algorithms for cvar optimization in mdps.
\newblock In \emph{Advances in neural information processing systems}, pp.\
  3509--3517, 2014.

\bibitem[Chow et~al.(2015{\natexlab{a}})Chow, Ghavamzadeh, Janson, and
  Pavone]{chow2015riskconstrained}
Chow, Y., Ghavamzadeh, M., Janson, L., and Pavone, M.
\newblock Risk-constrained reinforcement learning with percentile risk
  criteria, 2015{\natexlab{a}}.

\bibitem[Chow et~al.(2015{\natexlab{b}})Chow, Tamar, Mannor, and
  Pavone]{chow2015risk}
Chow, Y., Tamar, A., Mannor, S., and Pavone, M.
\newblock Risk-sensitive and robust decision-making: a cvar optimization
  approach.
\newblock In \emph{Advances in Neural Information Processing Systems}, pp.\
  1522--1530, 2015{\natexlab{b}}.

\bibitem[Coraluppi \& Marcus(1999)Coraluppi and Marcus]{coraluppi1999risk}
Coraluppi, S.~P. and Marcus, S.~I.
\newblock Risk-sensitive and minimax control of discrete-time, finite-state
  markov decision processes.
\newblock \emph{Automatica}, 35\penalty0 (2):\penalty0 301--309, 1999.

\bibitem[Depeweg et~al.(2018)Depeweg, Hernandez-Lobato, Doshi-Velez, and
  Udluft]{depeweg2018decomposition}
Depeweg, S., Hernandez-Lobato, J.-M., Doshi-Velez, F., and Udluft, S.
\newblock Decomposition of uncertainty in bayesian deep learning for efficient
  and risk-sensitive learning.
\newblock In \emph{International Conference on Machine Learning}, pp.\
  1192--1201, 2018.

\bibitem[Dimitrakakis(2011)]{dimitrakakis:mmbi:ewrl:2011}
Dimitrakakis, C.
\newblock Robust bayesian reinforcement learning through tight lower bounds.
\newblock In \emph{European Workshop on Reinforcement Learning (EWRL 2011)},
  pp.\  177--188, 2011.

\bibitem[Friedman \& Savage(1948)Friedman and Savage]{friedman1948uac}
Friedman, M. and Savage, L.~J.
\newblock {The Utility Analysis of Choices Involving Risk}.
\newblock \emph{The Journal of Political Economy}, 56\penalty0 (4):\penalty0
  279, 1948.

\bibitem[Ghavamzadeh \& Engel(2006)Ghavamzadeh and Engel]{ghavamzadeh:bpga}
Ghavamzadeh, M. and Engel, Y.
\newblock Bayesian policy gradient algorithms.
\newblock In \emph{{NIPS} 2006}, 2006.

\bibitem[Ghavamzadeh \& Engel(2007)Ghavamzadeh and
  Engel]{ghavamzadeh2007bayesian}
Ghavamzadeh, M. and Engel, Y.
\newblock Bayesian actor-critic algorithms.
\newblock In \emph{Proceedings of the 24th international conference on Machine
  learning}, pp.\  297--304. ACM, 2007.

\bibitem[Ghavamzadeh et~al.(2016)Ghavamzadeh, Engel, and
  Valko]{ghavamzadeh2016bayesian}
Ghavamzadeh, M., Engel, Y., and Valko, M.
\newblock Bayesian policy gradient and actor-critic algorithms.
\newblock \emph{The Journal of Machine Learning Research}, 17\penalty0
  (1):\penalty0 2319--2371, 2016.

\bibitem[Givan et~al.(2000)Givan, Leach, and Dean]{givan2000bounded}
Givan, R., Leach, S., and Dean, T.
\newblock Bounded-parameter markov decision processes.
\newblock \emph{Artificial Intelligence}, 122\penalty0 (1-2):\penalty0 71--109,
  2000.

\bibitem[Houthooft et~al.(2016)Houthooft, Chen, Duan, Schulman, De~Turck, and
  Abbeel]{houthooft2016vime}
Houthooft, R., Chen, X., Duan, Y., Schulman, J., De~Turck, F., and Abbeel, P.
\newblock Vime: Variational information maximizing exploration.
\newblock In \emph{Advances in Neural Information Processing Systems}, pp.\
  1109--1117, 2016.

\bibitem[Howard \& Matheson(1972)Howard and Matheson]{howard1972risk}
Howard, R.~A. and Matheson, J.~E.
\newblock Risk-sensitive markov decision processes.
\newblock \emph{Management science}, 18\penalty0 (7):\penalty0 356--369, 1972.

\bibitem[Leike et~al.(2017)Leike, Martic, Krakovna, Ortega, Everitt, Lefrancq,
  Orseau, and Legg]{leike2017ai}
Leike, J., Martic, M., Krakovna, V., Ortega, P.~A., Everitt, T., Lefrancq, A.,
  Orseau, L., and Legg, S.
\newblock Ai safety gridworlds.
\newblock \emph{arXiv preprint arXiv:1711.09883}, 2017.

\bibitem[Mannor et~al.(2012)Mannor, Mebel, and Xu]{mannor2012lightning}
Mannor, S., Mebel, O., and Xu, H.
\newblock Lightning does not strike twice: Robust mdps with coupled
  uncertainty.
\newblock \emph{arXiv preprint arXiv:1206.4643}, 2012.

\bibitem[Marcus et~al.(1997)Marcus, Fern{\'a}ndez-Gaucherand,
  Hern{\'a}ndez-Hernandez, Coraluppi, and Fard]{marcus1997risk}
Marcus, S.~I., Fern{\'a}ndez-Gaucherand, E., Hern{\'a}ndez-Hernandez, D.,
  Coraluppi, S., and Fard, P.
\newblock Risk sensitive markov decision processes.
\newblock In \emph{Systems and control in the twenty-first century}, pp.\
  263--279. Springer, 1997.

\bibitem[Markowitz(1952)]{markowitz1952portfolio}
Markowitz, H.
\newblock Portfolio selection.
\newblock \emph{The journal of finance}, 7\penalty0 (1):\penalty0 77--91, 1952.

\bibitem[Mihatsch \& Neuneier(2002)Mihatsch and Neuneier]{mihatsch2002risk}
Mihatsch, O. and Neuneier, R.
\newblock Risk-sensitive reinforcement learning.
\newblock \emph{Machine learning}, 49\penalty0 (2-3):\penalty0 267--290, 2002.

\bibitem[Mnih et~al.(2016)Mnih, Badia, Mirza, Graves, Lillicrap, Harley,
  Silver, and Kavukcuoglu]{mnih2016asynchronous}
Mnih, V., Badia, A.~P., Mirza, M., Graves, A., Lillicrap, T., Harley, T.,
  Silver, D., and Kavukcuoglu, K.
\newblock Asynchronous methods for deep reinforcement learning.
\newblock In \emph{International conference on machine learning}, pp.\
  1928--1937, 2016.

\bibitem[Murphy(2007)]{Mur07}
Murphy, K.
\newblock {Conjugate bayesian analysis of the gaussian distribution}.
\newblock Technical report, UBC, 2007.

\bibitem[O'Donoghue et~al.(2016)O'Donoghue, Munos, Kavukcuoglu, and
  Mnih]{o2016combining}
O'Donoghue, B., Munos, R., Kavukcuoglu, K., and Mnih, V.
\newblock Combining policy gradient and q-learning.
\newblock \emph{arXiv preprint arXiv:1611.01626}, 2016.

\bibitem[Pathak et~al.(2017)Pathak, Agrawal, Efros, and
  Darrell]{pathak2017curiosity}
Pathak, D., Agrawal, P., Efros, A.~A., and Darrell, T.
\newblock Curiosity-driven exploration by self-supervised prediction.
\newblock In \emph{Proceedings of the IEEE Conference on Computer Vision and
  Pattern Recognition Workshops}, pp.\  16--17, 2017.

\bibitem[Powell(2007)]{powell2007approximate}
Powell, W.~B.
\newblock \emph{Approximate Dynamic Programming: Solving the curses of
  dimensionality}, volume 703.
\newblock John Wiley \& Sons, 2007.

\bibitem[Rockafellar et~al.(2000)Rockafellar, Uryasev,
  et~al.]{rockafellar2000optimization}
Rockafellar, R.~T., Uryasev, S., et~al.
\newblock Optimization of conditional value-at-risk.
\newblock \emph{Journal of risk}, 2:\penalty0 21--42, 2000.

\bibitem[Sivaramakrishnan \& Stamicar(2017)Sivaramakrishnan and
  Stamicar]{sivara2017}
Sivaramakrishnan, K. and Stamicar, R.
\newblock A cvar scenario-based framework: Minimizing downside risk of
  multi-asset class portfolios.
\newblock \emph{The Journal of Portfolio Management}, 44:\penalty0 114--129, 12
  2017.
\newblock \doi{10.3905/jpm.2018.44.2.114}.

\bibitem[Stanko(2018)]{stanko2018distribuvcni}
Stanko, S.
\newblock Risk-averse distributional reinforcement learning, 2018.

\bibitem[Sutton et~al.(2000)Sutton, McAllester, Singh, and
  Mansour]{sutton2000policy}
Sutton, R.~S., McAllester, D.~A., Singh, S.~P., and Mansour, Y.
\newblock Policy gradient methods for reinforcement learning with function
  approximation.
\newblock In \emph{Advances in neural information processing systems}, pp.\
  1057--1063, 2000.

\bibitem[Tamar et~al.(2012)Tamar, Di~Castro, and Mannor]{tamar2012policy}
Tamar, A., Di~Castro, D., and Mannor, S.
\newblock Policy gradients with variance related risk criteria.
\newblock In \emph{Proceedings of the twenty-ninth international conference on
  machine learning}, pp.\  387--396, 2012.

\bibitem[Tamar et~al.(2014)Tamar, Mannor, and Xu]{tamar2014scaling}
Tamar, A., Mannor, S., and Xu, H.
\newblock Scaling up robust mdps using function approximation.
\newblock In \emph{International Conference on Machine Learning}, pp.\
  181--189, 2014.

\bibitem[Tamar et~al.(2015)Tamar, Glassner, and Mannor]{tamar2015optimizing}
Tamar, A., Glassner, Y., and Mannor, S.
\newblock Optimizing the cvar via sampling.
\newblock In \emph{Twenty-Ninth AAAI Conference on Artificial Intelligence},
  2015.

\bibitem[Williams(1992)]{williams1992simple}
Williams, R.~J.
\newblock Simple statistical gradient-following algorithms for connectionist
  reinforcement learning.
\newblock \emph{Machine learning}, 8\penalty0 (3-4):\penalty0 229--256, 1992.

\bibitem[Williams \& Peng(1991)Williams and Peng]{williams1991function}
Williams, R.~J. and Peng, J.
\newblock Function optimization using connectionist reinforcement learning
  algorithms.
\newblock \emph{Connection Science}, 3\penalty0 (3):\penalty0 241--268, 1991.

\end{thebibliography}
\bibliographystyle{icml2019}

\appendix
%\section{Algorithms}
%\label{sec:algorithms}
%\input{algorithms}
\section{Environment descriptions}
\subsection{Gridworld}
\label{sec:gridworld}
\label{sec:agw}
A visual representation of the environment can be seen in~\citet{leike2017ai}. The experiments are run with slightly different parameters, which are the following; Let $\gamma = 0.99$ be the discount factor, and $T = 100$ be the maximum allowed steps within an episode. The allowed actions correspond to the four cardinal directions $\{North, East, South, West\}$. The environment consists of a \emph{GOAL} tile, which terminates the episode and gives a reward of $r_g = 0.0$. The environment also has \emph{WATER} tiles which, upon entering, terminates the episode and gives a reward of $r_w = -10.0$. Every other action in any other state gives a reward of $r_0 = -0.1$. To ensure there is always epistemic uncertainty, we introduce another almost identical environment but with a goal at another location.

We measure performance in \emph{Regret} with respect to the optimal deterministic risk-neutral policy. The regret of an algorithm $\mathcal{A}$ with respect to the optimal deterministic risk-neutral policy is given by the following;

\begin{equation}
Regret(\mathcal{A}, \pi^*) = \int_{\MDP}\Big(V_\mu^{\pi^*} - \sum_{t=0}^T \gamma^t r_t\Big)d\mdp.
\end{equation}

We let our prior belief over the MDP transitions be $\mathcal{T}_\mdp^0(s'|s, a) = Dir(\alpha_0), \alpha_0 = 0.5$. We can then, given observed transitions in the environment, update $\mathcal{T}_\mdp^T(s'|s, a) = Dir(\frac{\alpha_0 + n_{s,a}^{s'}}{\sum_{s''} n_{s,a}^{s''}})$, that is, we count the number of transitions $(s,a) \rightarrow s'$ and $(s,a) \rightarrow \cdot$. From these counts we can compute $\alpha_t$ which is our posterior on the transition $(s,a) \rightarrow s'$ at time $t$. 

We also hold a belief over the reward matrices $\mathcal{R}_\mdp(s, a)$. A common prior is to use is a $NormalGamma$-prior~\cite{Mur07} over each $r_\mdp(s, a)$. The NormalGamma distribution is the conjugate prior to the $Normal$ distribution with unknown parameters $\mu, \lambda$. The model is given as $p(\mu, \lambda|D) = NG(\mu, \lambda|\mu_n, \kappa_n, \alpha_n, \beta_n)$. So our belief over rewards $\mathcal{R}_\mu(s,a) = NG(\mu, \lambda|\mu_n, \kappa_n, \alpha_n, \beta_n)$.

\subsection{Options}
\label{sec:option}
\label{sec:ao}
A common experiment~\cite{chow2014algorithms, tamar2012policy}, used to test risk-averse algorithms in a continuous environment is the \emph{Option Pricing} experiment. It can have many forms and classically we aim to minimise expected cost. In this work, we consider the reward perspective instead and the goal is to maximise the expected return. The environment is as follows; At any time $t$ the agent represents an investor with the opportunity of buying an option which gives an immediate reward of $r_t$. The reward of this option varies with time and with probability $p$, $r_{t+1} \leftarrow \gamma f_ur_t, $ and with probability $(1-p), r_{t+1} \leftarrow \gamma f_dr_t$. The goal is then for the agent to chose a suitable time to stop. Upon choosing to buy the option, the episode ends and the agent returns to $x_0$. Every time step the agent decides to wait will also incur a small negative reward $p_h$ which represents the opportunity cost for not using its resources. If the agent chooses to wait until the horizon $T$, it is then forced to take the option for its current value.

We use the following parameters in our experiment;
$x_0 = [1; 0], p_h = -0.1, T = 20, \gamma = 0.95, f_u = 2.0, f_d = 0.5, p = [0.45; 0.65; 0.85], K=5.0$. 

Algorithm~\ref{alg:papg} requires us to be able to sample MDPs from the environment and to be able to do rollouts in them. There are no obvious candidates of priors over continuous state-space transitions and reward functions. In this work, we chose to use Gaussian Processes as priors over functions.
We use multiple Gaussian Processes to maintain our belief over transition kernels and reward functions. We update our belief with transitions and reward from the true underlying models. We use a Dirichlet prior over models and update it when we receive information on which model we acted in.

Given data from rollouts sampled from our belief, we can use any standard Policy gradient-like algorithm to update our policy $\pol_{\theta_{t}}$. In this paper, we chose to focus on an episodic Policy gradient algorithm similar to the \emph{REINFORCE} framework. We parametrise our policy by a one hidden layer neural network with parameters $\theta$.

\end{document}